\documentclass{sig-alternate}

\usepackage{amsmath} 
\usepackage{dsfont}
\usepackage{amssymb} 
\usepackage{color}
\usepackage[bookmarks=false]{hyperref}

\newcommand{\Rbb}{\mathbb{R}}

\graphicspath{{figures/}}

\newif\ifcomment
\newif\iftodo

\newtheorem{remark}{Remark}

\definecolor{innerboxcolor}{rgb}{.9,.95,1}
\definecolor{outerlinecolor}{rgb}{.6,0,.2}
\definecolor{todocolor}{rgb}{.1, 1, .1}
\newcommand{\red}[1]{\textcolor{red}{#1}}

\newcommand{\todo}[1]{\iftodo\fcolorbox{outerlinecolor}{todocolor}{
    \begin{minipage}{.9\columnwidth}
      \red{\bf TODO:} {#1}
  \end{minipage}} \\
  \fi
}

\DeclareMathOperator*{\Prob}{\mathds{P}}
\DeclareMathOperator*{\card}{card}

\newcommand \Ncal{\mathcal{N}}
\newcommand \Scal{\mathcal{S}}
\newcommand \Lcal{\mathcal{L}}

\newcommand \begMat{\left( \begin{array}}
\newcommand \enMat{\end{array} \right)}

\newcommand \begDet{\left| \begin{array}}
\newcommand \enDet{\end{array} \right|}

\newcommand \ind[1]{\mathds{1}_{\{#1\}}}
\newcommand \eqv{\Leftrightarrow}

\newcommand \rel[2]{#1 \rightarrow #2}

\newcommand \relpn[2]{#1 \overset{p^{(n)}}{\rightarrow} #2}

\newcommand \nTraces N
\newcommand \nTracePoints J
\newcommand \nSGChanges K

\newcommand \TMM T

\newcommand \stopgo {\emph{Stop-\&-Go} filter}
\newcommand \hmm {Markov model}
\newcommand \gmrf {GMRF}

\newcommand \pathTT[1] {Z^{(#1)}}
\newcommand \linkTT[1] {Z^{#1}}

\newcommand \condPathTT[2] {\pathTT{#1}_{\left|#2\right.}}

\newenvironment{denselist}{\begin{list}{$\bullet$}%
{\setlength{\itemsep}{0ex} \setlength{\topsep}{0ex}
\setlength{\parsep}{0pt} \setlength{\itemindent}{0pt}
\setlength{\leftmargin}{1em}
\setlength{\partopsep}{0pt}}}%
{\end{list}}

\begin{document}


\title{Arriving on time: estimating travel time distributions on large-scale road networks}

\numberofauthors{4} 
%

\author{
Timothy Hunter\qquad Aude Hofleitner\qquad Jack Reilly\qquad Walid Krichene\qquad J{\'e}r{\^o}me Thai\\
       \affaddr{UC Berkeley}\\
       \email{\{tjhunter, aude.hofleitner, jackdreilly, walid, jerome.thai\}@berkeley.edu}
\and
\alignauthor
Anastasios Kouvelas\\
       \email{kouvelas@berkeley.edu}
\alignauthor
Pieter Abbeel\\
       \email{pabbeel@berkeley.edu}
\alignauthor
Alexandre Bayen\\
       \email{bayen@berkeley.edu}
}

\date{\today}

\maketitle

\begin{abstract}
Most optimal routing problems focus on minimizing travel time or distance traveled. Oftentimes, a more useful objective is to maximize the probability of on-time arrival, which requires statistical distributions of travel times, rather than just mean values. We propose a method to estimate travel time distributions on large-scale road networks, using probe vehicle data collected from GPS. We present a framework that works with large input of data, and scales linearly with the size of the network. Leveraging the planar topology of the graph, the method computes efficiently the time correlations between neighboring streets. First, raw probe vehicle traces are compressed into pairs of travel times and number of stops for each traversed road segment using a `stop-and-go' algorithm developed for this work. The compressed data is then used as input for training a path travel time model, which couples a Markov model along with a Gaussian Markov random field. Finally, scalable inference algorithms are developed for obtaining path travel time distributions from the composite MM-GMRF model.
We illustrate the accuracy and scalability of our model on a 505,000 road link network spanning the San Francisco Bay Area.
\end{abstract}





\section{Introduction}\label{sec:introduction}

A common problem in trip planning is to make a given deadline, for example
arriving at the airport within 45 minutes. Most routing services available today minimize the expected travel time, but do not provide the most \emph{reliable} route, which accounts for the variability in travel
times. Given a time budget, a routing service
should provide the route with highest probability of on-time arrival, as posed in stochastic on-time arrival (SOTA) routing~\cite{Samaranayake2011stochastic}. 
Such an algorithm requires the estimation of the \emph{statistical distributions} of travel times, or at least of their means and variances, as done in~\cite{limpractical}. Today, only few traffic information platforms are available for the arterial network (the state of the art for highway networks is more advanced) and they do not provide the \emph{statistical distributions} of travel times. The main contribution of the article is precisely to addresses this gap: we present a scalable algorithm
for learning path travel time distributions on the entire road network using probe vehicle data.

The increasing penetration rate of probe vehicles provides a promising source of data to learn and estimate travel time distributions in arterial networks. At present, there are two general trends in estimation of travel times using this probe data. One trend, from kinematic wave theory (see~\cite{zheng_itsc_2010,hofleitner_isttt_2010}), derives analytical probability distributions of travel times and infer their parameters with probe vehicle data. These approaches are computationally intensive, which limits their applicability for large scale networks. The other trend, seen in large-scale navigation systems such as Google Maps, provides coarser information, such as expected travel time, but can scale to world-sized traffic networks. 

We bridge the two trends by creating a travel time estimator that (i) provides full probability distributions for arbitrary paths in real-time (sub-second), (ii) works on networks the size of large cities (and perhaps larger) (iii) and accepts an arbitrary amount of input probe data. The model uses a data-driven model which is able to leverage physical insight from traffic flow research. Data-driven models, using dynamic Bayesian networks~\cite{hofleitner_tits_2012}, nearest neighbors~\cite{tiesyte2008similarity} or Gaussian models~\cite{westgate2011travel} show the potential of such methods to make accurate predictions when large amounts of data are available.

\begin{figure}
\centering \includegraphics[width=.8\columnwidth]{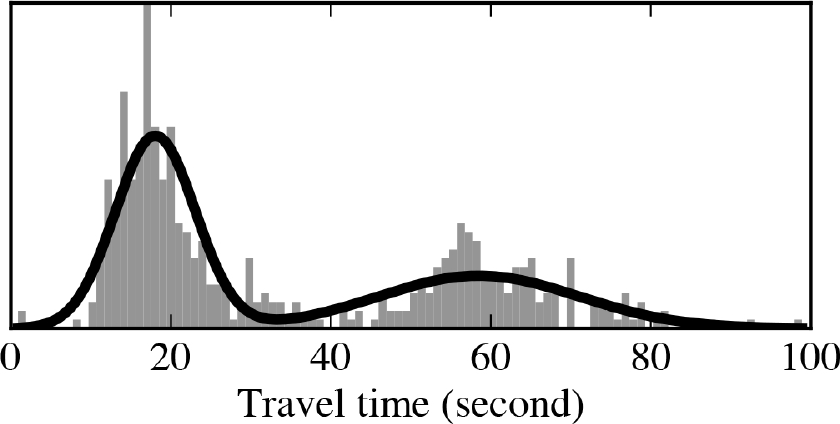}
\vspace*{-.2cm}
\caption{Histogram of travel times collected on a link fit (solid line) using a mixture of Gaussian distributions.}
\label{fig:example_bimodal_link}
\vspace*{-.5cm}
\end{figure}

The main physical insights modeled in the article are described in the following paragraphs.
First, arterial traffic is characterized by important travel time variability amongst users of the network (Figure~\ref{fig:example_bimodal_link}). This variability is mainly due to the presence of traffic lights and 
other impediments such as pedestrian crossings and garbage trucks which cause a fraction of the vehicles to stop while others do not. Arterial traffic research~\cite{hofleitner_isttt_2010} suggests that the detection of stops explains most of the variability in the travel time distribution and underline the multi-modality of the distributions.

Second, the number of stops on a trajectory exhibits strong spatial
and temporal correlations. Traffic lights on major streets may be 
synchronized to create ``green waves'': a vehicle which does not
stop on a link is likely to not stop on the subsequent link. A different
vehicle arriving 10 seconds later may hit the red light on the first
link and have a high probability to stop on
the subsequent link. This phenomenon is analyzed in~\cite{Ramezani2012markov} using a Markov model with two modes (``slow'' and ``fast'').

Third, besides the the number of stops, travel times may be correlated for the following reasons: (i) the behavior of individual drivers may be different: some car may
travel notably faster than some others, (ii) congestion propagates on the network, making neighboring links  likely to have similar congestion levels. If a driver experiences a longer than usual travel time on a link because of heavy traffic, he / she can will likely experience heavy traffic on the subsequent links.

\begin{figure}
\centering \includegraphics[width=1\columnwidth]{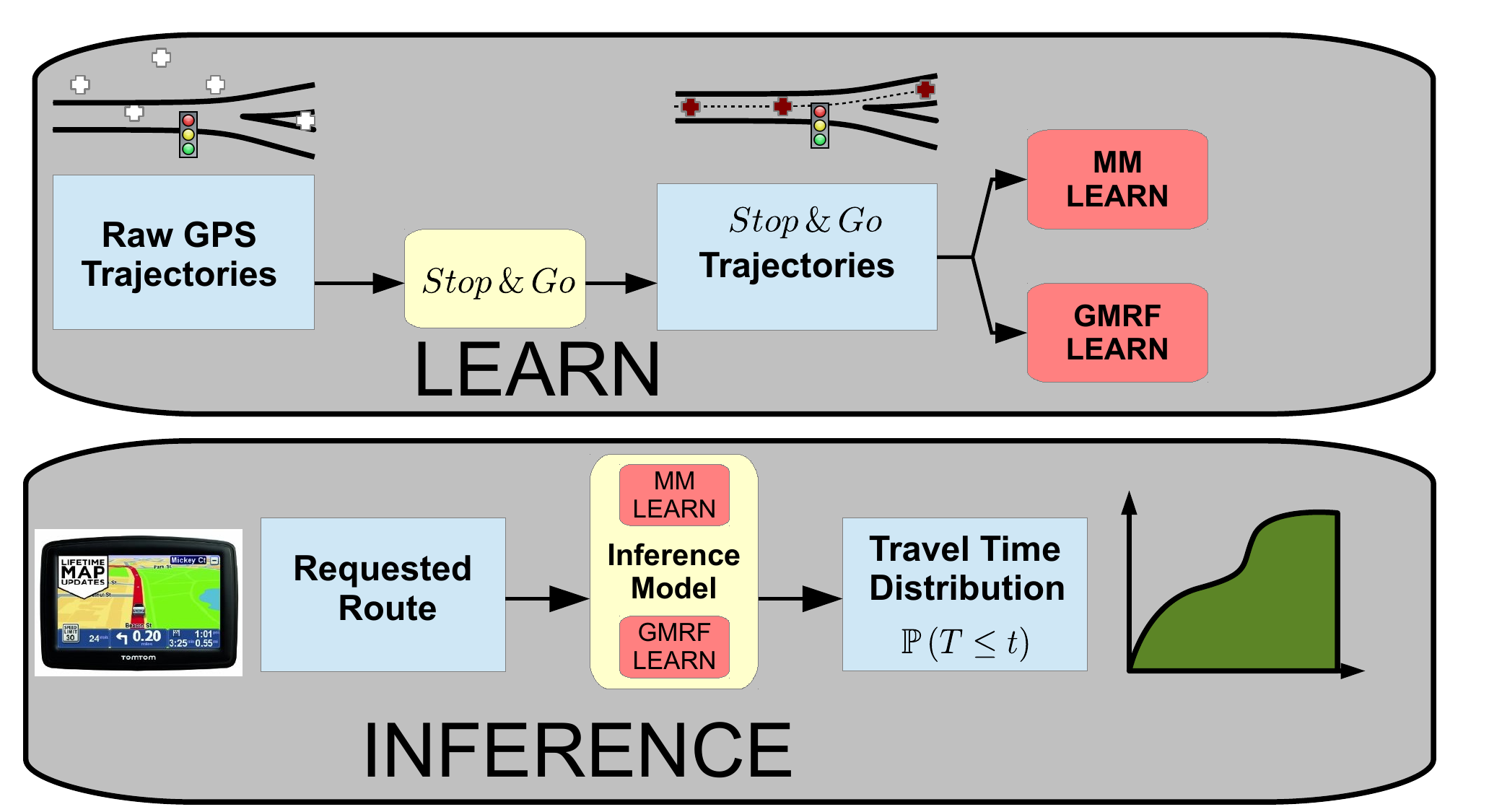}
\caption{Pipeline of the travel-time estimation model. The learning runs offline and finds optimal parameters for the
\hmm{} and \gmrf{} models. The inference runs online and
uses the learned parameters to produce travel time distributions on
input paths.}
\label{fig:pipeline} 
\end{figure}

We leverage these insight to develop the traffic estimation algorithm presented in this article: an end-to-end, scalable model for inferring path travel
time distributions, referred to as a ``pipeline'' (see Figure~\ref{fig:pipeline}). It consists of a \textit{learning algorithm} and an \textit{inference algorithm}.

The\textit{ learning algorithm  }consists in the following steps.\\
$\bullet$ Section~\ref{sec:stop-and-go}: a \stopgo{} algorithm detects the number of stops on a link and compresses the GPS traces\footnote{Before feeding raw GPS points to the \stopgo{}, each coordinate is mapped to a link and an offset distance from the beginning of the link. We use an efficient path-matching and path-inference algorithm developed in~\cite{hunter12wafr}.} into values of travel times on traversed links and corresponding number of stops. \\ 
$\bullet$ Section~\ref{sec:hmm}: a \textit{Markov model} (MM) captures the correlations of stopping / not
stopping for consecutive links. It characterizes the probability to stop / not stop on a link given that the vehicle stopped or did not stop on the previous link traversed. The \stopgo{} produces a set of labeled data
to train the Markov model. \\
$\bullet$ Section~\ref{sec:gmrf-model}: a \textit{Gaussian Markov Random
Field} (\gmrf{}) captures the correlations in travel times between
neighboring links, given the number of stops on the links. 
There is a significant body of prior work in the field of learning with graphical models~\cite{jordan_introduction_1999, jordan_learning_1999},
especially for learning problems on sparse \gmrf{}s~\cite{gu2007learning,malioutov2008low}.
Most of these algorithms do not scale linearly with respect to the
dimension of the data, and are unsuitable for very large problems (hundreds
of thousands of variables). In particular, it becomes practically impossible 
to store the entire covariance matrix, so even classical sub-gradient
methods such as~\cite{duchi08projected} would require careful engineering.

%
We exploit the (near) planar structure
of the underlying graphical model to more efficiently obtain (approximate)
algorithms that scale \emph{linearly} with the size of the network.  Our algorithm leverages efficient algorithms to compute the Cholesky decomposition 
of the adjacency matrix of planar graphs~\cite{chen2008algorithm}.
Our results can be extended to other physical systems with
local correlations.

After the learning, we proceed to the \textit{inference algorithm}: we compute the travel time distributions for arbitrary paths in the network (Section~\ref{sec:inference}). While exact inference on this model
is intractable due to the number of possible states, we exploit
the underlying structure of the graphical model and use a specialized
sampling method to obtain an efficient inference algorithm.

Section~\ref{sec:evaluation} illustrates the accuracy and scalability of our model on a 505,000 road link network spanning the San Francisco Bay Area.

\todo{Add something about website here}

%
%

Our code, as well as a showcase of the model running on San Francisco, is available at \url{http://traffic.berkeley.edu/navigateSF}.


\section{\secit{Stop-\&-Go} model for vehicles trajectories}\label{sec:stop-and-go}

The stops due to traffic signals and other factors (double parking, garbage trucks and so on) represent one of the main source of variability in urban travel times. More generally, consider that a link can have $m$ different discrete states. For a vehicle traveling on link $l$, the \emph{state} $s_l \in \{0, m - 1\}$ of the trajectory is defined as the number of stops on the link. The following algorithm estimates the number of stops given a set of noisy GPS samples from the trajectory on link $l$. We consider a generic trajectory on a generic link and drop indices referring to the trajectory and link for notation simplicity.

The trajectory of the vehicle is represented by an offset function
$T: [0, \tau] \rightarrow \Rbb_+$, representing the distance from the beginning of the link to the location of the vehicle at time $t$. The noisy GPS observations are defined by the times $0 = t_0 < \dots < t_J \leq \tau$, and the corresponding offsets $x_j = T(t_j) + \epsilon_j$, where $\epsilon_j \sim \Ncal(0, \sigma^2)$ are independent and identically distributed zero mean Gaussian random variables representing the GPS noise. 


We process the observations $(t_j, x_j)_{j = 0, \dots, J}$ to obtain \emph{stop and go} trajectories of the probe vehicles: the trajectory of a vehicle alternates between phases of \emph{Stop} during which the velocity of the vehicle is zero and \emph{Go}  during which the vehicle travels at positive speed. 
The number of \emph{Stop} phases represents the state of the trajectory. We assume that the sampling frequency is high enough that the speed between successive observations $(t_j, x_j)$ and $(t_{j+1}, x_{j+1})$ is constant\footnote{The assumption is further justified by traffic modeling~\cite{hofleitner_isttt_2010} which commonly assumes that each \emph{Go} phase has constant speed} and denoted $v_j$. Note that speeds are rarely provided by GPS devices or are too noisy to be valuable for estimation.

Maximizing the log-likelihood of the observations is equivalent to solving the following optimization problem
$$
\underset{(v_j)_{j \in \{0, \ldots, J-1\} }}{\mathbf{minimize}} \frac{1}{2} \sum_{j=0}^{J-1}  \left(x_{j+1} - x_0 -\sum_{j'=0}^j v_{j'}(t_{j'+1} - t_{j'})\right)^2
$$
This is a typical least-square optimization problem, which we conveniently write in matrix form as:
\begin{equation}
\underset{v}{\mathbf{minimize}} \frac{1}{2} || A v \, - \, b||_2^2,
\label{eq:stop-go:ls}
\end{equation}
with the notation $v = (v_j)_{j \in \{0, \ldots, J-1\} }$, $b = (x_{j} - x_0)_{j \in \{1, \ldots, J\} }$ and $A$ is the lower triangular matrix whose entry on line $i \in \{0,\ldots, J-1\}$ and column $k \in \{0,\ldots, i-1\}$ is given by $t_{k+1} - t_k$.

\begin{figure}
\def\svgwidth{.6\columnwidth} 
\centering
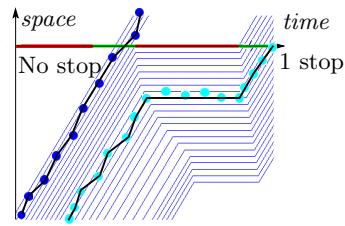
\vspace*{-.6cm}
\caption{Example of trajectory estimation}
\label{fig:trajectory-estimation-example}
\end{figure}

To detect the \emph{Stop} phases, we add a $l_1$ regularization term, with parameter $\lambda$ to Problem~\eqref{eq:stop-go:ls}. The resulting optimization problem is known as the LASSO~\cite{tibshirani1996regression} and reads
\begin{equation}
\underset{v}{\mathbf{minimize}} \frac{1}{2}|| A v \, - \, b||_2^2 + \lambda ||v||_1,
\label{eq:stop-go:lasso}
\end{equation}
The solution of~\eqref{eq:stop-go:lasso} is typically sparse~\cite{tibshirani1996regression}, which means that there is a limited number of non-zero entries, corresponding to the \emph{Go} phases of the trajectory. The solution is used to compute the number of \emph{Stops}.
Figure~\ref{fig:trajectory-estimation-example} illustrates the result of the trajectory estimation and Stop detection algorithm.

\begin{remark}[Data compression]
In our dataset\footnote{The number of GPS points per link depends on the level of congestion (vehicles spend a longer amount of time on each link), average length of the links of the network and sampling frequency of the probe vehicles.}, the average number of GPS points sent by a vehicle on a link is 9.6. The algorithm reduces the GPS trace to: entrance time in the link, travel time, number of stops. The amount of data to be processed by the subsequent algorithms is reduced by a factor of almost 10, which is crucial for large scale applications which process large amounts of historical data. 
\end{remark}

\begin{remark}[Fixing $\lambda$]
In~\eqref{eq:stop-go:lasso}, $\lambda$ acts as a trade-off between the sparsity of the solution and the fit to the observations. Cross-validation is not appropriate in our setting to fix this parameter as it would require one to decimate the trajectory and use some observations for the learning and others for the validation. Instead,  $\lambda$ is chosen by computing the \emph{Bayesian Information Criterion} (BIC), using~\cite{zou2007degrees} to estimate the  number of degrees of freedom. A LARS implementation~\cite{efron2004least} allows efficient computation to choose $\lambda$ and compute the corresponding solution.
\end{remark}

\section{Travel time model}\label{sec:graph-model}

We develop a travel time model which exploits the compressed data returned by the \stopgo{} (number of stops and travel time experienced per link). The model captures the state transition probabilities: the probability of the number of stop on a link given the number of stops on the previous link of the trajectory. It also models the correlations of travel times for neighboring links given their state (number of stops).
The travel time model is a combination of two models:\\
$\bullet$ A directed Markov model of discrete state random variables gives the joint probability distribution of the link states $\Prob_{\theta}(\{S_{l} \})$.\\
$\bullet$ A Gaussian Markov random field gives the joint distribution of the travel times $\Prob_{\theta}(\{Y_{l, s} \}, l \in \Lcal, s \in \{0, \dots, m - 1\})$.

\begin{figure}[h]
\centering
\includegraphics[width=8cm]{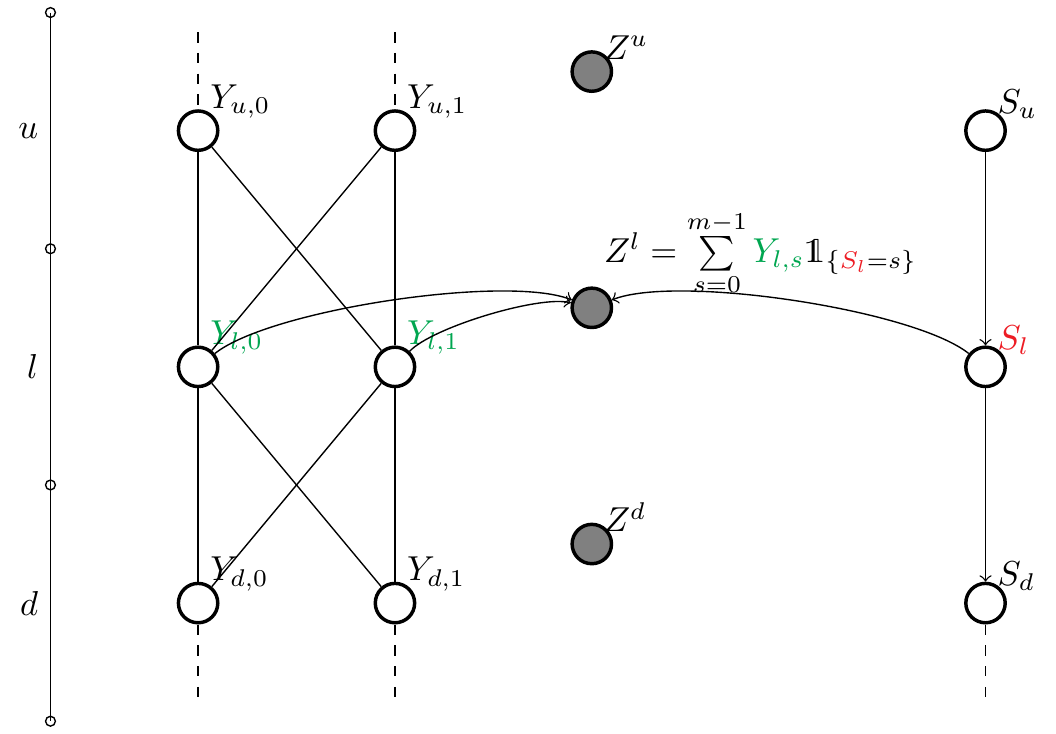}
\caption{Graphical model of the dependency of link travel time $\linkTT{l}$ on the state variable (e.g. number of stops) $S_{l}$ and the conditional travel times $Y_{l, s}$. Here we consider the subgraph consisting of a link $l$, upstream link $u$ and downstream link $d$ on a given path.}
\label{fig:graph_model}
\end{figure}

Figure~\ref{fig:graph_model} presents the graphical model that encodes the dependencies between the link travel time $\linkTT{l}$, the link states $S_{l}$ and the conditional travel times $Y_{l, s}$.  The total travel time experienced by a vehicle on link $l$, is $\linkTT{l} = \sum_{s = 0}^{m - 1} Y_{l, s} \ind{S_{l} = s}$. The left portion of the figure shows a subset of the GMRF of travel time variables, and the right portion shows a subset of the Markov chain of states. 

The graphical model shows that conditioning on the states experienced along a path allows one to compute the path travel time by summing over the corresponding variables in the \gmrf{}. Further, when one conditions on the link travel times $\linkTT{l}$, then the two models become independent, which allows one to learn the models parameters separately. The rest of this section details the modeling and learning of the two models.

\subsection{Markov model for state transitions}\label{sec:hmm}

We consider the state variables $\{S_l\}_{l, \in \Lcal}$. Each variable $S_{l}$ has support $\{0, \dots, m - 1\}$. 

Given the path $p = (l_0, \dots, l_M)$ of a vehicle, the variables $\{S_{l_i}\}_{i \in \{1, \dots, M\} }$ have a Markov property, i.e. given the state of the \emph{upstream link} $l_{i-1}$, the conditional state $(S_{l_i} | S_{l_{i-1}})$ is independent of the state of other upstream links:


\begin{equation}
\label{eq:state_markov_prop}
\Prob(S_{l_i} | \{S_{l_{i-1}}, \dots, S_{l_0}\}) = \Prob(S_{l_i} | S_{l_{i-1}})
\end{equation}

We parametrize the model using an initial probability vector
\begin{equation}
\label{eq:init_proba_vector}
\pi^{l}_{s} = \Prob(S_{l} = s)
\end{equation}
and a transition probability matrix
\begin{equation}
\label{eq:transition_matrix}
T^{\rel{u}{l}}_{s_u, s_l} = \Prob(S_{l} = s_l | S_{u} = s_u)
\end{equation}
here $T^{\rel{u}{l}}_{s_u, s_l}$ is the probability that $l$ is in state $s_l$ given that the upstream link $u$ is in state $s_u$.





We learn the initial probability vector $\pi^{l}$ and the transition probability matrices $T^{\rel u l}$ of the Markov Chain by maximizing the likelihood of observing  $(s^{(n)}, y^{(n)})$. The log-likelihood is given by
\[
\sum_{n = 1}^M \left( \log(\pi^{l^{(n)}_0}_{s_0^{(n)}}) + \sum_{\relpn u l} \log (T^{\rel u l}_{s_u^{(n)}, s_l^{(n)}}) \right)
\]




The parameters that maximize the log-likelihood are:
\begin{align*}
\pi^{l}_s  \propto \bar{\pi}^{l}_s \quad \textrm{and} \quad
T^{\rel u l}_{s_u, s_l}  \propto \bar{T}^{\rel u l}_{s_u, s_l}
\end{align*}
where $\bar{\pi}^{l}_s$ are the empirical counts of initial states and $\bar{T}^{\rel u l}_{s_u, s_l}$ are the empirical transition counts.
This solution corresponds to transitions and initial probabilities that are consistent with the empirical counts of initial states and transitions.




\subsection{\gmrf{} model for travel time distributions}\label{sec:gmrf-model}

We now present the model that describes the correlations between the
travel time variables. We assume that the random variables $\left(Y_{l,s}\right)_{l\in\mathcal{L},s\in\mathcal{S}_{l}}$ are Gaussian\footnotemark, and they can be stacked into one multivariate Gaussian variable $Y\sim\mathcal{N}\left(\mu,\Sigma\right)$ of size $m\card(\Lcal)$, mean $\mu$ and covariance $\Sigma$. Recall
that $Y_{l,s}$ is the travel time on link $l$ conditioned on state $s$, where $s\in\{0, \dots, m-1\}$. This travel time is a Gaussian random
variable with mean $\mu_{l,s}$ and variance $\sigma_{l,s}$.

\footnotetext{While Gaussian travel times can theoretically predict negative travel time, in practice, these probabilities are virtually zero, as validated in Section~\ref{sec:evaluation}.}

From the factorization property given by the Hammersley-Clifford
Theorem, it is well known that the \emph{precision matrix} $S=\Sigma^{-1}$ encodes the \emph{conditional
dependencies} between the variables. 
Since a link is assumed to be conditionally correlated only with its neighbors, the precision matrix is very
sparse. Furthermore, this sparsity pattern is of particular
interest: its pattern is that of a graph which is nearly planar. We take advantage of this structure to devise efficient algorithms that (1) estimate the precision
matrix given some observations, and (2), infer the covariance between
any pair of variables.

As mentioned earlier, we have a set of $N$ trajectories obtained from GPS data. After map-matching and trajectory reconstruction (Section~\ref{sec:stop-and-go}), the set of observed trajectories $\{W_{p}:p=1,\cdots,N\}$ are sequences of observed states
and variables (travel time)%
: 
\begin{eqnarray*}
W_{p} & = & \left(l_{1},s_{1}\right)\,\left(l_{2},s_{2}\right)\,\,\cdots\left(l_{M_{p}},s_{M_{p}}\right)\\
y^{p} & = & \left(y^{p}_{l,s}\right)_{(l,s)\in W_{p}}
\end{eqnarray*}

In our notations, $y^{p}$ is an observation of the random vector $Y_{p}=\left(Y_{i}\right)_{i\in W_{p}}$, which is a $M_{p}$-dimensional marginal (or subset) of the full distribution $Y$. Hence $Y_{p}$ is also a multivariate Gaussian with mean $\mu_{(W_{p})}$ and covariance $\Sigma_{(W_{p})}$ obtained by dropping the irrelevant variables (the variables that one wants to marginalize out) from the mean vector $\mu$ and the covariance matrix $\Sigma$. Its likelihood is thus the likelihood under the marginal distribution:
\begin{equation}
\begin{array}{l}
\text{log } p\left(y^{p};\mu_{(W_{p})},\Sigma_{(W_{p})}\right) = \\
C -\frac{1}{2}\log\left|\Sigma_{(W_{p})}\right|
 -\frac{1}{2}\left(y^{p}-\mu_{(W_{p})}\right)^{T}\Sigma_{(W_{p})}^{-1}\left(y^{p}-\mu_{(W_{p})}\right),
\end{array}
\end{equation}
where $C$ is a constant which does not depend on the parameters of the model.

The problem of estimating the parameters of the model $\theta=(\mu,\Sigma)$ from the i.i.d. set of observations $\mathcal{D}=\{W_{p},y^{p}:p=1,\cdots,N\}$ consists in finding the set of parameters $\theta^{\star}$ that maximize the sum of the likelihoods of each of the observations
:
\begin{equation}
\label{eq:gmrf-II}
l\left(\theta|\mathcal{D}\right)=\sum^{N}_{p=1}\text{log }p\left(y^{p};\mu_{(W_{p})},\Sigma_{(W_{p})}\right)
\end{equation}

Unfortunately, the problem
of maximizing~\eqref{eq:gmrf-II} is in general not
convex and may have multiple local minima since we have only partially observed variables $y^{p}$.
A popular strategy in this case is to \emph{complete} the vector (by
computing the most likely completion given the observed variables). This algorithm is called the \textit{Expectation-Maximization} 
(EM) procedure. In our case, the EM procedure is not a good fit for two reasons:
\begin{denselist}
\item Since we observe only a small fraction of the values of each vector,
the vast majority of the values we would use for learning would be
sampled values, which would make the convergence rate dramatically
slow.
\item The data completion step would create a complete $n-$size sample
for each of our observation, thus our complexity for the data completion
step would be $\mathcal{O}\left(N n\right)$, which is too large to
be practical.
\end{denselist}
Instead, we solve a related problem by computing sufficient statistics
from all the observations. Consider the simpler scenario in which
all data has been observed, and denote the empirical covariance
matrix by $\tilde{\Sigma}$.
The maximum likelihood problem to find the most likely precision matrix
is then equivalent to: 
\begin{equation}
\underset{S}{\mathbf{minimize}}\,\,-\log\left|S\right|+\text{Tr}\left(S\tilde{\Sigma}\right)\label{eq:gmrf-maxll-emp}
\end{equation}
under the structured sparsity constraints $S_{uv}=0\,\,\forall\,\left(u,v\right)\notin\mathcal{E}$.
The objective is not defined when $S$ is not positive definite, so
the constraint that $S$ is positive definite is implicit. A key point
to notice is that the objective only depends on a restricted subset
of terms of the covariance matrix: 
\[
\text{Tr}\left(S\tilde{\Sigma}\right)=\sum_{\left(u,v\right)\in\mathcal{E}}S_{uv}\tilde{\Sigma}_{uv}
\]

This observation motivates the following approach: instead of considering
the individual likelihoods of each observation individually, we consider
the covariance that would be produced if all the observations were
aggregated into a single covariance matrix. This approach discards
some information, for example the fact that some variables are more
often seen than others. However, it lets us solve the full covariance
Problem~\eqref{eq:gmrf-maxll-emp} using partial observations. Indeed,
all we need to do is estimate the values of the coefficients $\tilde{\Sigma}_{uv}$
for $\left(u,v\right)$ a non-zero in the precision matrix. We present
one way to estimate these coefficients.

Let $N_{i}$ be the number of observations of the variable $Y_{i}$: $N_{i} = \text{card}\left(\{p:i\in W_{p}\}\right)$. Combining all the observations that come across $Y_{i}$, we can approximate
the mean of any function $f\left(Y_{i}\right)$ by some empirical
mean, using the $N_{i}$ samples: 
\begin{equation}
\mathbb{E}_{i}\left[f\left(Y_{i}\right)\right]=\frac{1}{N_{i}}\sum_{p:i\in W_{p}}f\left(y^{p}_{i}\right)
\end{equation}

Similarly, defining the number of observations of both $Y_{i}$ and $Y_{j}$: 
$N_{i\cap j}=\text{card}\left(\left\{ p\,:\,i\in W_{p}\text{ and }j\in W_{p}\right\} \right)$, we can approximate the mean of any function $f\left(Y_{i},Y_{j}\right)$
of $Y_{i},Y_{j}$, using the set of observations that span both variables
$Y_{i}$ and $Y_{j}$:

\begin{equation}
\mathbb{E}_{i\cap j}\left[f\left(Y_{i},Y_{j}\right)\right]=\frac{1}{N_{i\cap j}}\sum_{p\,:\,i,j\in W_{p}}f\left(y_{i}^{p},y_{j}^{p}\right)
\end{equation}

Using this notation, the empirical mean is $\hat{\mu}_{i}=\mathbb{E}_{i}\left[Y_{i}\right]$. Call $\hat{\Sigma}$ the \emph{partial empirical covariance matrix}
(PECM): 
\[
\hat{\Sigma}_{ij}=\begin{cases}
\mathbb{E}_{i\cap j}\left[Y_{i}Y_{j}\right]-\mathbb{E}_{i}\left[Y_{i}\right]\mathbb{E}_{j}\left[Y_{j}\right] & \text{if }\left(i,j\right)\in\mathcal{E}\\
0 & \text{otherwise}
\end{cases}
\]

Using this PECM as a proxy for the real covariance matrix, one can
then estimate the most likely GMRF by solving the following problem:
\begin{equation}
\underset{S}{\mathbf{minimize}}\,\,-\log\left|S\right|+\left\langle S,\hat{\Sigma}\right\rangle \label{eq:gmrf-maxll-gecm}
\end{equation}

Note that this definition is asymptotically consistent: in the limit,
when an infinite number of observations are gathered ($N_{ij}\rightarrow\infty$),
the PECM will converge towards the true covariance: indeed $\hat{\Sigma}_{ij}\rightarrow\mathbb{E}\left[Y_{i}Y_{j}\right]-\mathbb{E}\left[Y_{i}\right]\mathbb{E}\left[Y_{j}\right]$
and for all $S$, $\left\langle S,\hat{\Sigma}\right\rangle \rightarrow\left\langle S,\mathbb{E}\left[YY^{T}\right]-\mathbb{E}\left[Y\right]\mathbb{E}\left[Y\right]^{T}\right\rangle $. 

Unfortunately, the problem is not convex because $\hat{\Sigma}$ is not necessarily positive
semi-definite (even if the limit is), since the variables are only partially observed. For instance, if we have a partially observed bivariate Gaussian variable $X$: $\left(10,10\right)$, $\left(-10,-10\right)$, $\left(1,\star\right)$, $\left(-1,\star\right)$, $\left(\star,1\right)$, $\left(\star,-1\right)$, the empirical covariance matrix $\hat{\Sigma}$ has diagonal entries $(51, 51)$ and off-diagonal entries $(100, 100)$. Its eigenvalues are $-49, 151$ hence it is not definite positive.

There is a number of ways to correct this. The simplest we found is
to scale all the coefficients so that they have the same variance:
\[
\hat{\Sigma}_{ij}=\begin{cases}
\alpha_{ij}\mathbb{E}_{i\cap j}\left[Y_{i}Y_{j}\right]-\mathbb{E}_{i}\left[Y_{i}\right]\mathbb{E}_{j}\left[Y_{j}\right] & \text{if }N_{ij}>0\\
0 & \text{otherwise}
\end{cases}
\]
\[
\alpha_{ij}=\sqrt{\frac{\mathbb{E}_{i}\left[Y_{i}^{2}\right]\mathbb{E}_{j}\left[Y_{j}^{2}\right]}{\mathbb{E}_{i\cap j}\left[Y_{i}^{2}\right]\mathbb{E}_{i\cap j}\left[Y_{j}^{2}\right]}}
\]

This transformation has the advantage of being local and easy to compute.
This is why it is completed by an increase of the diagonal coefficients
by some factor of the identity matrix.

Another problem is due to the relative imbalance between the distributions
of samples: cars travel much more on some roads than others. This
means that some edges may be much better estimated than some others,
but this confidence does not appear in the PECM. In practice, we found that \emph{pruning} the edges with too few observations improved the results


\section{Inference}\label{sec:inference}

Given the model parameters, the inference task consists in computing the probability distribution $\Prob(Z^{(p)} \le t)$ of the total travel time along a fixed path $(p) = (p_0, \dots, p_I)$. The path travel time (i.e.\ total travel time along the path) is given by
\[
\pathTT{p}
= \sum_{s \in \Scal} \ind{S_{(p)} = s} \condPathTT{p}{s} 
\]
where $\Scal = \{1, \dots, m\}^{I}$ is the set of possible path states, and $\condPathTT{p}{s}$ is the path travel time given the path state $s$, and is given by $\condPathTT{p}{s}  = \sum_{l \in p} Y_{l, s_l} = e(p, s)^T Y$. Here, $e(p, s)$ is a binary vector that selects the appropriate entries in the vector of travel times $Y$ (corresponding to path $p$ with states $s$):
\[
e(p, s)_{(l, s'_l)} = 1 \eqv l \in p \text{ and } s'_l = s_l
\]
This vector $e(p,s)$ is very sparse and has precisely $I$ non-zero entries.
Using the law of total probability we have:
\begin{equation}
\label{eqn:total-prob}
\Prob\left({\pathTT{p}}\right)=\sum_{s \in \Scal} \Prob(S = s)\Prob\left({\condPathTT{p}{s}}\right)
\end{equation}
The variable $\condPathTT{p}{s} = e(p, s)^TY$ is a linear combination of the multivariate variable $Y$, and so 
is also normally distributed:
\begin{equation}
\label{eqn:cond-prob}
\condPathTT{p}{s}\sim\mathcal{N}\left(e(p, s)^{T}\mu,e(p, s)^{T}\Sigma e(p, s)\right)
\end{equation}

The marginal distribution of travel times along a path is a mixture of univariate Gaussian distributions. There is
however two problems for this algorithm to be practical. \emph{(i)} The mixture from Equation~\eqref{eqn:total-prob} contains
a term for each possible combination of states, and has size $m^I$. Enumerating all the terms is impossible for
moderately large lengths of paths.
\emph{(ii)} In order to compute the variance of each distribution of the mixture, one needs to estimate the covariance
matrix $\Sigma$. However storing (or computing) the complete covariance matrix is prohibitively expensive with millions
of variables.

We find tractable solutions for \emph{(i)} by using a sampling method on the Markov model to choose a tractable number of states, and \emph{(ii)} by using a random projection method to construct a low-rank approximation of the covariance matrix $\Sigma$. Note that the mean of the complete distribution can be computed exactly in $O(Im^2)$ time, using a dynamic programming algorithm. Thus, our model is also applicable to situations that do not require the full distribution. 
We do not present this algorithm further due to space limitations.

\paragraph*{Gibbs sampling}

The sequence of state variables $\{S_l\}_l$ for a given path form a Markov chain, with initial probability $\pi^{0}$ and transition probability matrices $T^{\rel{i-1}{i}}$, which are given by the trained Markov model in Section~\ref{sec:hmm}. We can sample from the Markov chain by first sampling $s_{0}$ from the categorical distribution with parameter $\pi^{0}$, then sequentially sample $s_{i}$ from the conditional distribution of $S_{i}|S_{i-1}$.

Using this procedure, we generate $K$ samples $\hat{s}^1, \dots, \hat{s}^K$ from
$\Scal$, the set of possible states. The complete (exponential) distribution can be approximated with the 
empirical distribution:

\begin{equation}
\label{eq:approx_total_prob}
\Prob\left({\pathTT{p}}\right)=\sum_s \hat{w}_{p, s} \Prob\left({\condPathTT{p}{s}}\right)
\end{equation}

in which each weight $\Prob$ is the fraction of samples corresponding to the state $s$:
\begin{equation}
\label{eq:approx_weights}
\hat{w}_{p, s} = \frac{1}{K} \card \{k | \hat{s}^k = s\}
\end{equation}
The sum~\eqref{eq:approx_total_prob} contains at most $K$ terms, since at most $K$ approximate weights are positive, 
and converges towards the true distribution of states as the number of samples increases. In
practice, numerical experiments showed that it was only necessary to sample a relatively small number of states. 
Section~\ref{sec:evaluation} includes a numerical analysis and evaluation of the sampling method.

\paragraph*{Low rank covariance approximation}


Compute the variances of each sequence of variables:
\[
{e}\left({s}\right)^{T}\Sigma \, e \left({s}\right)
\]
with ${s}$ being a valid sequence of variables in the graph
of variables.

Using the full covariance matrix $\Sigma$ to estimate the covariance
of each path $\sigma\left(s\right)^{2}=e\left(p,s\right)^{T}\Sigma e\left(p,s\right)$
is impractical for two reasons: as mentioned before, we cannot expect
to compute and access the full covariance matrix, and also the sum
$e\left(p,s\right)^{T}\Sigma e\left(p,s\right)$ sums $I^{2}$ elements
from the covariance matrix. Since we do not need to know the variance terms
with full precision, an approximation strategy using random projections
is appropriate. More specifically, we use the following result from~\cite{achlioptas2001database}:

\emph{Given some fixed vectors $v_{1}\cdots v_{n}\in\mathbb{R}^{d}$
and $\epsilon>0$, let $R\in\mathbb{R}^{k\times d}$ be a random matrix
with random Bernoulli entries $\pm1/\sqrt{k}$ and with $k\geq24\epsilon^{-2}\log n$.
Then with probability at least $1-1/n$:
\[
\left(1-\epsilon\right)\left\Vert v_{i}\right\Vert ^{2}\leq\left\Vert Rv_{i}\right\Vert ^{2}\leq\left(1+\epsilon\right)\left\Vert v_{i}\right\Vert ^{2}
\]
}

Call $R$ such a matrix. Consider the Cholesky decomposition of the
precision matrix, $S=LL^{T}$. Then 
\begin{eqnarray*}
\sigma\left(s\right)^{2} & = & e\left(p,s\right)^{T}\Sigma e\left(p,s\right)\\
 & = & e\left(p,s\right)^{T}L^{-T}L^{-1}e\left(p,s\right)\\
 & = & \left\Vert L^{-1}e\left(p,s\right)\right\Vert ^{2}
\end{eqnarray*}
From the following lemma, we can approximate this norm:
\begin{equation}
\hat{\sigma}^{2}\left(s\right) = \left\Vert Q^{T}e\left(p,s\right)\right\Vert ^{2} = \left\Vert \sum_{i=1\cdots I}Q_{\left(l_{i},s_{i}\right)}\right\Vert ^{2}
\end{equation}
with $Q=L^{-1}R$ and $R$ defined as obtained from the lemma above.
Computing the approximate variance $\hat{\sigma}^{2}$ requires the
addition of $I$ vectors of size $k$. In practice, this summing operation
is vectorized and very fast.

This method assumes we can efficiently compute the Cholesky factorization,
and that inversion operation $L^{-1}x$ is efficient as well. In our
case, the graph of the GMRF is nearly planar. Some very efficient
algorithms exist that compute the Cholesky factorization in near-linear
time (\cite{chen2008algorithm}
). In practice, computing the $Q$ matrix is very fast.

\section{Evaluation}\label{sec:evaluation}

The article presents an algorithm to turn GPS traces from a small fraction of the vehicles traveling on the road network into valuable traffic information to develop large scale traffic information platforms and optimal and robust routing capabilities. The value of the model depends crucially on the quality of the estimates of point to point travel time distribution. Another key feature of the algorithm is its computational complexity and its ability to scale on large road networks. The large number of road segments and intersections leads to high-dimensional problems for any network of reasonable size. Moreover, the estimate for travel time distributions between any two points on the network needs to be computed in real-time. It would not be acceptable to wait more than a few seconds to get the results.
This section first analyzes the quality of the learning and the estimation of the model. Then, we study the computational complexity of the algorithm and its ability to scale on large networks. In particular, some aspects of the algorithm arise as trade-offs between the computational complexity and the desired precision in the learning and real-time inference.

The validation results are based on anonymized GPS traces provided by a GPS data aggregator. We consider an arterial network in the Bay Area of San Francisco, CA with 506,585 links.
The algorithm processes 426 million GPS points, aggregated from 2,640,319 individual trajectories. Each trajectory is less than 20 minutes long for privacy reasons. 

\subsection{Travel time distribution}\label{sub:eval-tt}

The full validation of the performance of the algorithm requires the observation of the travel time of every vehicle on every link of the network. This mode of validation is unfortunately not available for any reasonably sized network. We validate the learning capabilities of the algorithm using data which was not used to train the \hmm{} \gmrf{}. On this validation dataset, we perform two types of validation: (i) a path-level validation (a limited set of individual paths are evaluated) and (ii) a network-level validation (metrics taken over the entire  validation dataset).

\subsubsection*{Comparison models}

In this Section, the results of the model are compared to simpler models which arise as special cases of the model presented in the article. We introduce these models and denominations, which we use throughout the evaluation.\\
$\bullet$ \emph{One mode independent}: The travel time distribution on each segment is Gaussian. The travel time on distinct segments are independent random variables.\\
$\bullet$ \emph{One mode}: The travel time distribution on the network is a multi-variate Gaussian (one dimension per link). In the precision matrix, element $(i,j) $ may be non-zero if $i$ and $j$ map to neighboring links in the road network.\\
$\bullet$ \emph{Multi-modal independent}: Same as the MM-GMRF, excepted that the covariance matrix of the multi-variate Gaussian is diagonal, imposing that given the mode, the travel times on different links of the network are independent.\\
The model developed in this article is referred to as the \emph{MM-GMRF} model.

\subsubsection*{Path validation}
Most probe vehicles have different paths throughout the road network. Among the trajectories of the vehicles in the validation set, we select a set of paths for which a large enough number of vehicles has traveled to perform statistical validation of the distribution of travel times. We impose a minimum length (set to 150 m in the numerical experiments) on these paths to ensure that we validate the learning of the spatial distributions of the modes (Section~\ref{sec:hmm}) and the spatial correlations between each mode (Section~\ref{sec:gmrf-model}). The paths are selected for having the largest amount of validation data.

\begin{figure}
\centering
\includegraphics[width=\linewidth]{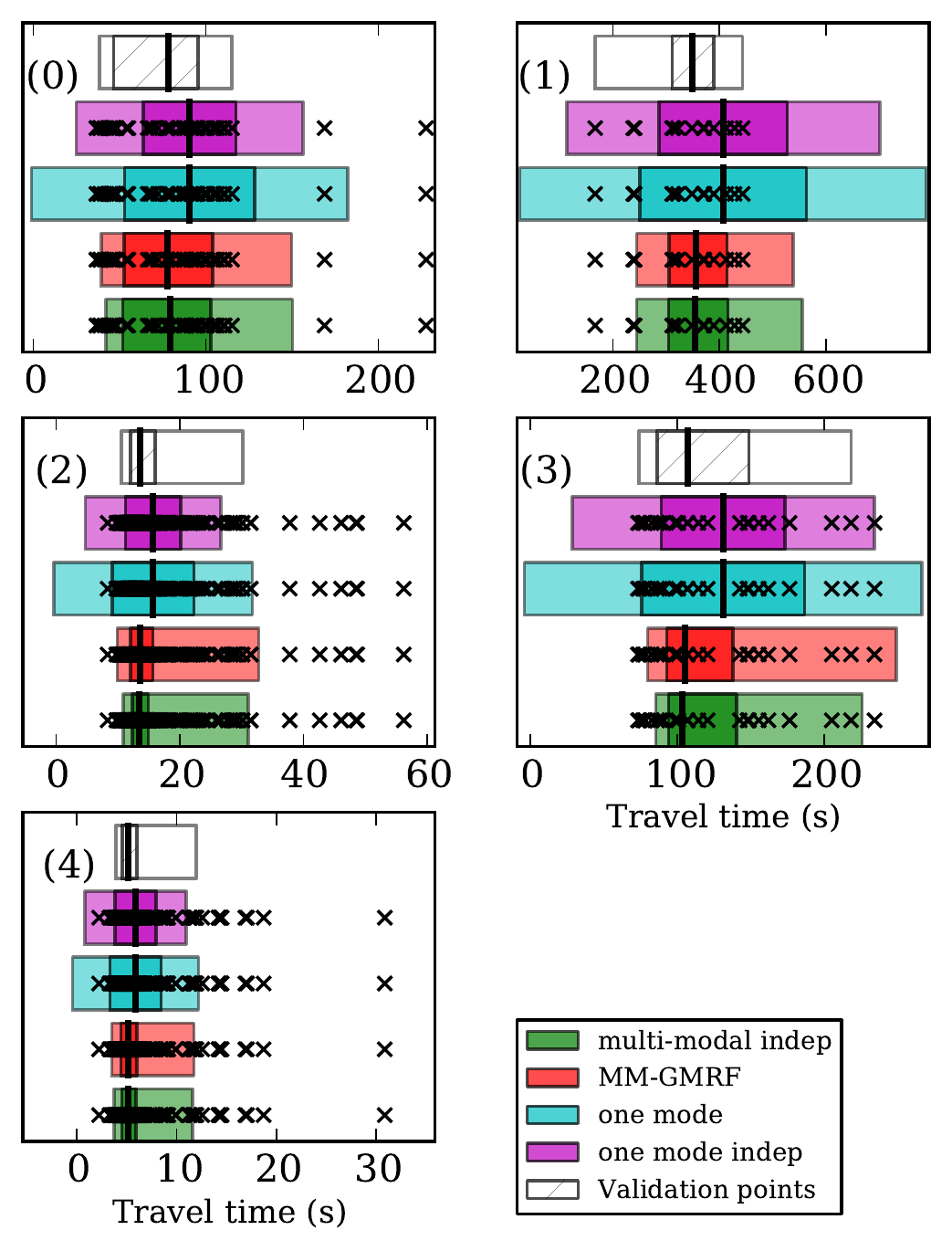}
\caption{50\% and 90\% confidence intervals computed by the different models and compared to the validation travel times on the selected paths.}
\label{fig:eval:scatter}
\end{figure}

For each selected path, the box plots on Figure~\ref{fig:eval:scatter} compare the 50\% and 90\% confidence intervals of the validation data collected on the path (top box) with the intervals computed by the different models (\textit{Multi-modal independent}, \textit{MM-GMRF} (our model), \textit{One mode} and \textit{One mode independent}, from top to bottom). We also display the median travel times as a vertical black line, both for the validation data and the different models. Scatter crosses, representing the validation travel times, are super-imposed to the results of each model to improve visualization.

We notice a significant difference in the results between the uni-modal models (\textit{One mode} and \textit{One mode independent}) and the multi-modal models (\textit{Multi-modal independent}, \textit{MM-GMRF}). The uni-modal models tend to over-estimate both the median and the variance of travel times. These models cannot account for the difference of travel times due to stops on trajectories, which is one of the main features of arterial traffic~\cite{hofleitner_isttt_2010}. The over-estimated variance illustrates why it is important to incorporate the variability of travel times due to stops in the structure of the model. On the other side, the multi-modal models are able to capture the features of the distribution fairly accurately. The differences in accuracy between the \textit{Multi-modal independent} model and the MM-GMRF model (which takes into account correlations in the Gaussian distribution) are not significant, even though the model with correlations estimates the variability slightly more accurately. It seems that capturing the variability of travel times due to stop is the most important feature of the model.

\begin{figure}
\centering
\includegraphics[width=\linewidth]{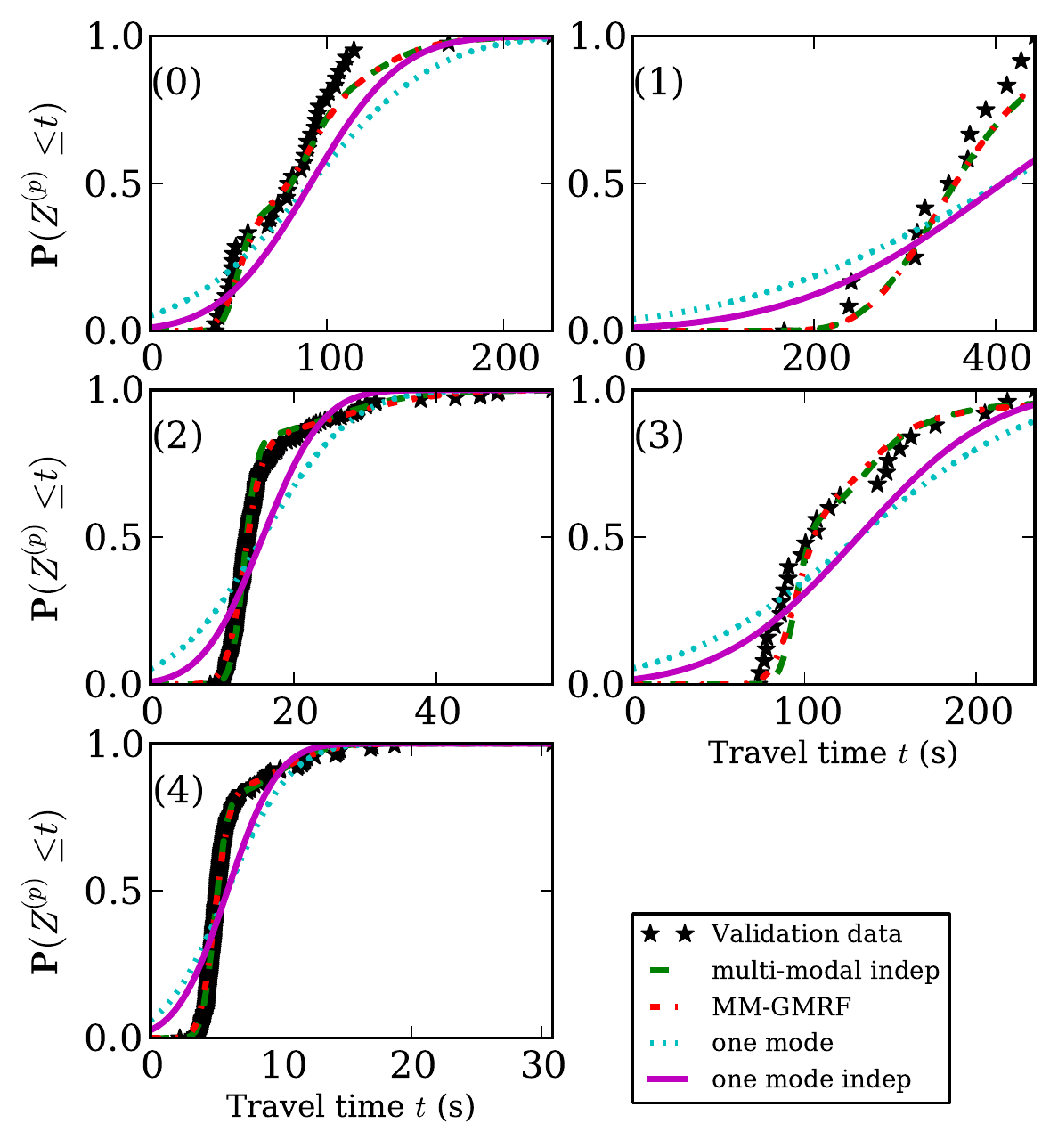}
\caption{Cumulative distribution of travel times computed by the different models and compared to the validation travel times received on the selected paths.}
\label{fig:eval:cdf}
\end{figure}

In Figure~\ref{fig:eval:cdf}, we display the cumulative distribution of the validation data and the cumulative distribution of each model for the same paths as for Figure~\ref{fig:eval:scatter} (displayed in the same order). The figure displays more precisely the difference in the estimation accuracy of the different models. As seen in Figure~\ref{fig:eval:scatter}, the multi-modal models are more accurate than their uni-modal counterparts. 

\subsubsection*{Network scale validation}
Most points in the observation dataset represent different paths for the probe vehicles. For this reason, the distributions cannot be compared directly. Instead, we compute the log-likelihood of each validation path and analyze the quality of the travel time bounds provided by the distribution for each path.

\begin{figure}
\centering
\includegraphics[width=8cm]{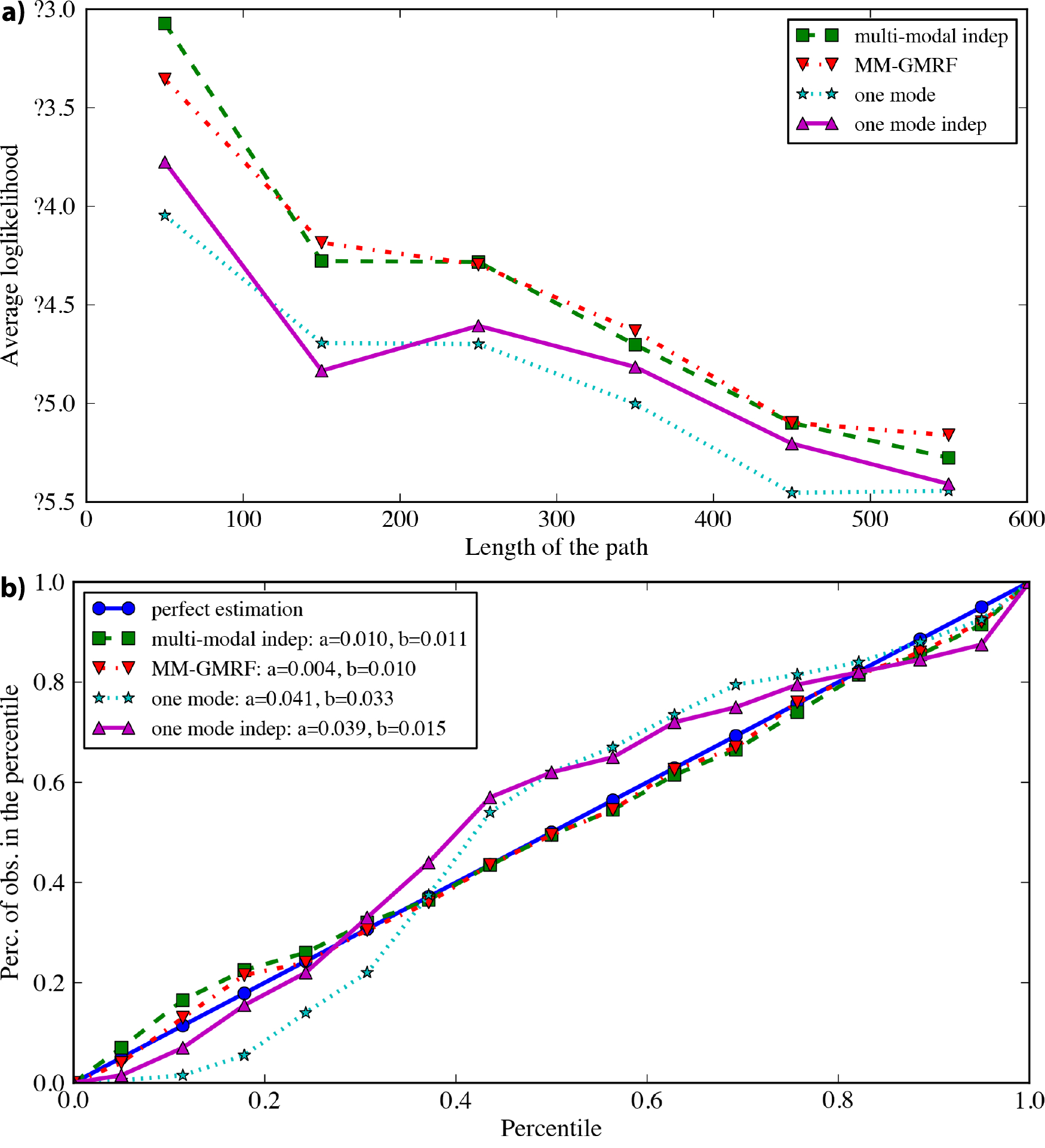}
\caption{a) Average log-likelihood of the validation paths (by path length), for each model. b) Validation of the distribution percentiles for each model.}
\label{fig:eval:llinterval}
\end{figure}

Figure~\ref{fig:eval:llinterval} a) displays the average likelihood of the validation paths computed by the different models. The figure also analyzes how the path length influences the results. There are two motivations for doing so: (i) the length of the path influences the support of the distribution (longer paths are expected to have a larger support) which may affect the likelihood and (ii) the different models may perform differently on different lengths as they do not take into account spatial dependencies in a similar way.

As was expected from the analysis of Figures~\ref{fig:eval:scatter} and~\ref{fig:eval:cdf}, the multi-modal models perform better than their uni-modal counterparts. Compared to the path validation results, the figure shows more significantly the effect of correlations. The figure shows slight improvements for the multi-modal model which takes into account the correlations. Surprisingly, the contrary is true for the uni-modal models. We also notice that the likelihood decreases with the length of the path, as we were expecting.

Figure~\ref{fig:eval:llinterval} b) analyzes the quality of the travel time distribution computed on the network. For that, we use a \emph{p-p plot} (or percentile-percentile plot) which assesses how much each learned distribution matches the validation data. To each path $p$ in the validation dataset corresponds an inverse cumulative distribution $\Prob_p^{-1}$ (computed from the trained model) and a travel time observation $z^p$. A point $(\alpha, \beta)$ on the curve corresponds to having $\beta$ percent of the validation points such that $z^p \leq P_p^{-1}(\alpha)$. If the estimation was perfect, there would be exactly $\alpha$\% of the data points in the percentile $\alpha$. To quantify how much each model deviates from perfect estimation, we display two metrics denoted $a$ (above) and $b$ (below). Let $f$ correspond to the p-p curve of a model, the corresponding metrics are computed as follows:
$$
a = \int_0^1 \max(f(\alpha) - \alpha, 0) \, d\alpha \, , \, b= \int_0^1 \max(\alpha - f(\alpha), 0) \, d\alpha
$$

These values provide insight on the quality of the fit of the model. For example, a model with a large $a$-value tends to overestimate travel times. Similarly, a model with a large $b$-value tends to underestimate travel times. 
Both uni-modal models have large $a$-values. The large variance estimated by the models (already noticed in Figure~\ref{fig:eval:scatter}) to account for the variability of travel times leads to non-negligible probability densities for small travel times which are not physically possible. Compared to the likelihood validation of Figure~\ref{fig:eval:llinterval} a), the p-p plot analyzes the quality of fit for different percentiles of the distribution. In particular, we notice that the effect of capturing the correlations in the multi-variate model mostly affects the estimation of the low and high percentiles in the distribution. We expect that this is due to the fact that correlations accounts for the impacts of slow vs. fast drivers or congested vs. lest congested conditions.


\subsection{Sampling}
\label{sub:sampling-evaluation}

In this section, we discuss numerical results regarding the quality of approximate inference using the Gibbs sampling method, on a fixed path on the network.
%
%
%

\begin{figure}[h]
\centering
\includegraphics[width = .24\textwidth]{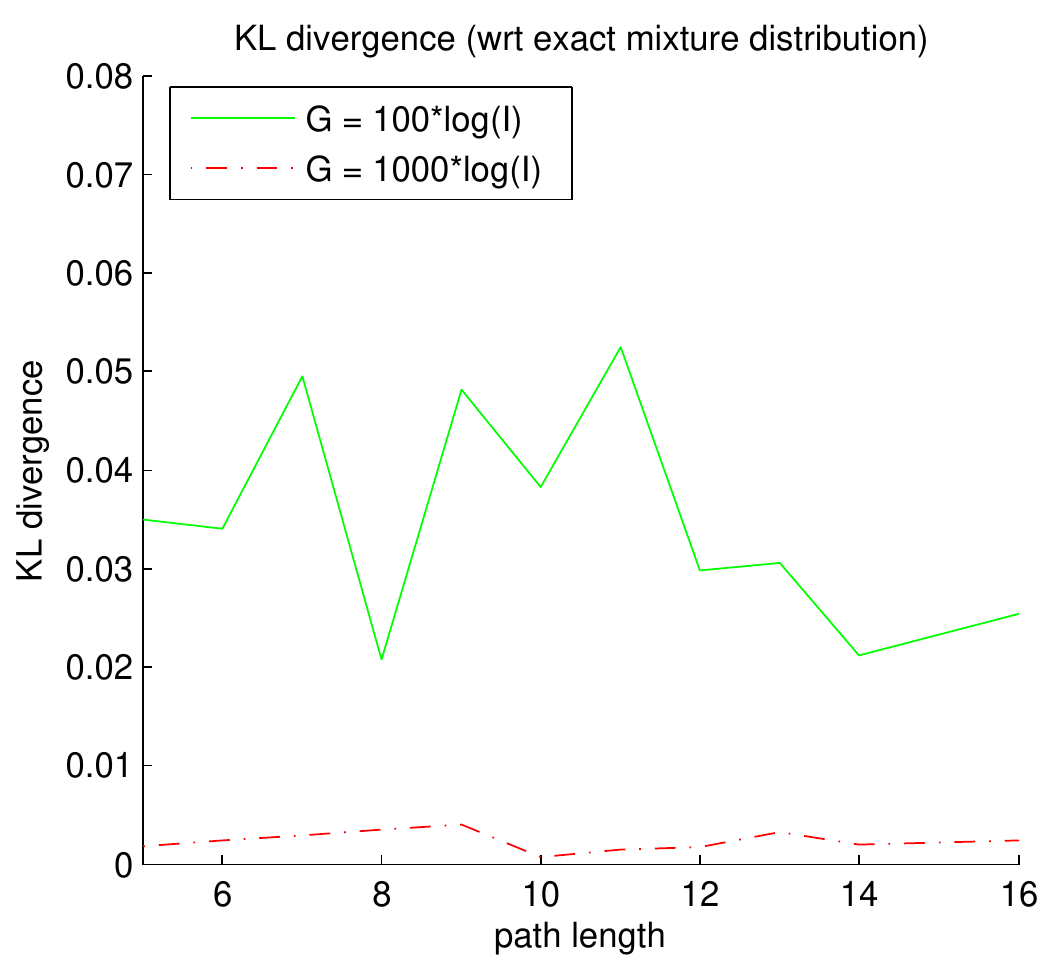}%
\includegraphics[width = .24\textwidth]{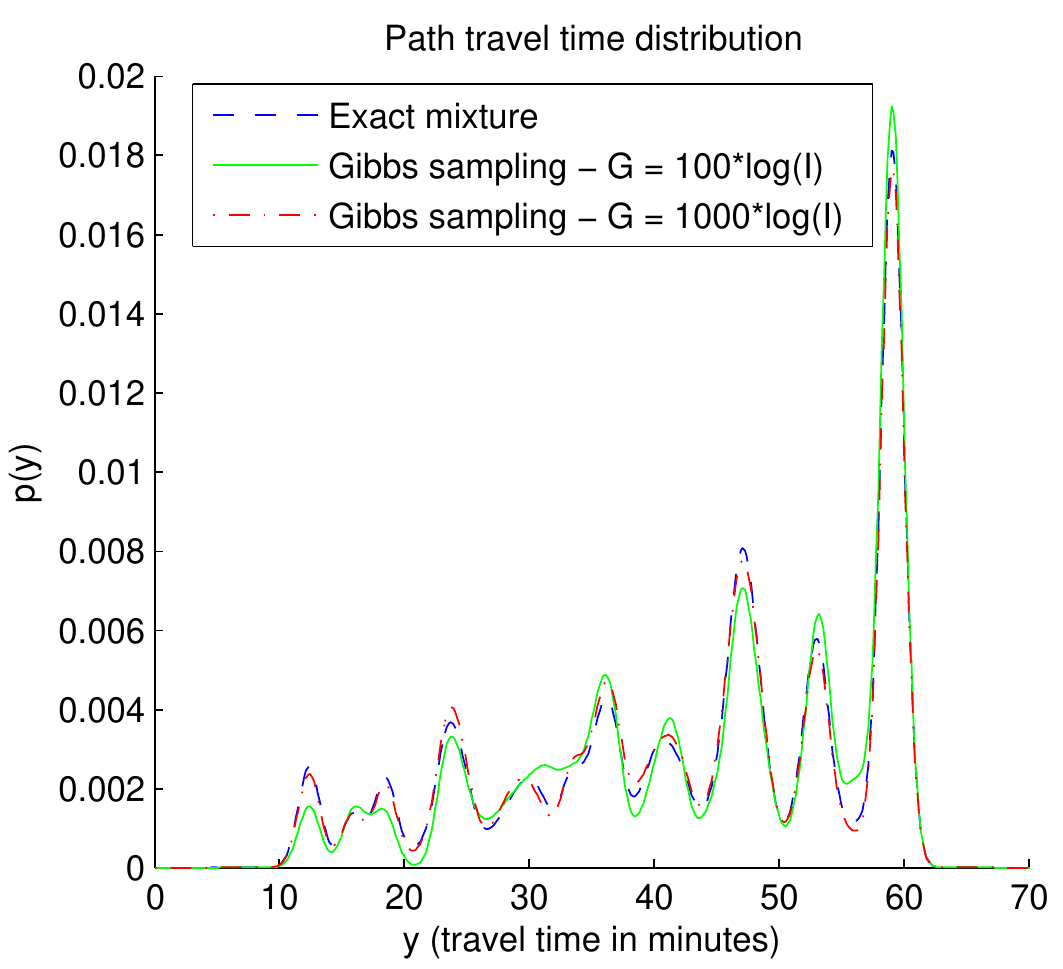}
\caption{KL-divergence between the approximate distribution and the exact mixture distribution, as a function of the path length (left), and example distributions for a path of length $I = 17$ (right).}
\label{fig:kl_div}
\end{figure}

Figure~\ref{fig:kl_div} shows the Kullback-Leibler divergence of the approximate distributions, with respect to the exact distribution. We compare two runs of the sampling, with respective sizes $G = 100 \log(I)$ and $G = 1000 \log(I)$ where $I$ is the length of the path in links. The divergence measures the similarity between two distributions. As can be seen, even a
small number of samples (relative to the total size of the mixture) leads to a very close approximation.

\begin{figure}[h]
\centering
\includegraphics[width = .5\textwidth]{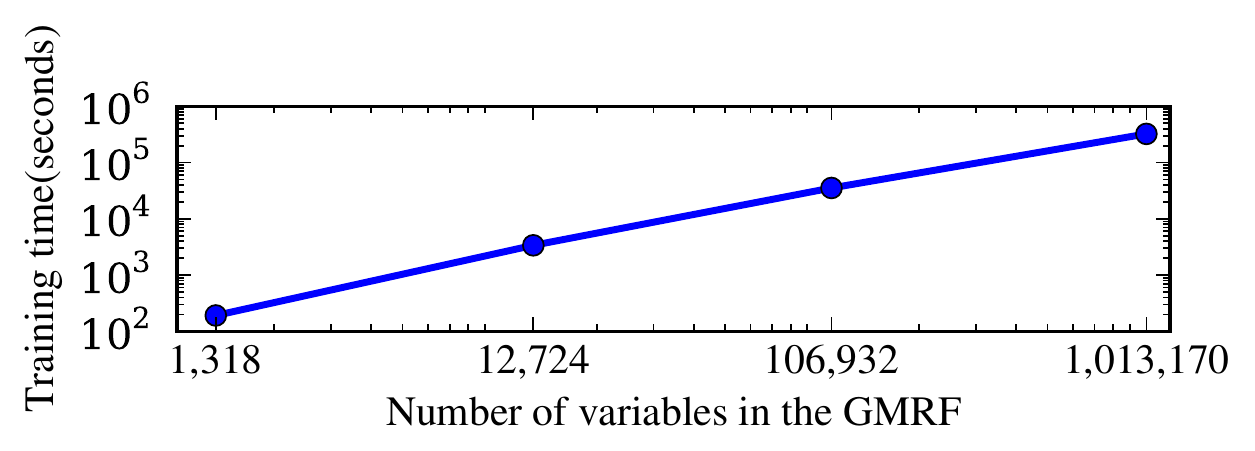}%
\caption{Log-log plot of the training time, as a function of the size of the GMRF.}
\label{fig:perf_gmrf}
\end{figure}

\subsection{Scaling}\label{sub:eval-scaling}

In this section, we discuss the scalability aspects of the learning algorithm (it is clear from the discussion in Section~\ref{sec:gmrf-model} that the inference
 is independent from the size of the network). We ran the learning for networks defined by different bounding boxes. The bounding boxes
were adjusted so that the number of links in each subnetworks had different orders of magnitude. The longest step by far is the training
of the GMRF. We report in Figure~\ref{fig:perf_gmrf} the training time for different networks, all other parameters being equal. As one can see, the training
time increases linearly with the number of variables of the GMRF over a large range of network sizes. The graphs associated to each GMRF are
extremely sparse: the average vertex degree of the largest graph is 9.46.

\section{Conclusions}\label{sec:conclusions}

The state of the art for travel time estimation has focused on either precise and computationally intensive physical models,
or large scale, data-driven approaches.
We have presented a novel algorithm for travel time estimation that aims at combining the best of both
worlds, by combining physical insights with some scalable algorithms. We model the variability of travel times due to stops at intersections using a \stopgo ~ (to detect stops) and a \hmm{} to learn the spatial dependencies between stop locations. We also take into account the spatio-temporal correlations of travel times due to driving behavior or congestion, using a Gaussian Markov Random Fields. In particular, we present a 
highly scalable algorithm to train and perform inference on Gaussian Markov Random Fields, when applied on 
geographs.

We analyze the accuracy of the model using probe vehicle data collected over the Bay Area of San Francisco, CA. The results underline the importance to take into account the multi-modality of travel times in arterial networks due to the presence of traffic signals. The quality of the results we obtain are competitive with the state of the 
state of the art in traffic, and also highlight the good scalability of our algorithm.

\bibliographystyle{abbrv}
\bibliography{bib}

\balancecolumns
\end{document}